\documentclass[12pt,twoside]{article}
\usepackage{latexsym}
\def\,{\mskip 3mu} \def\>{\mskip 4mu plus 2mu minus 4mu} \def\;{\mskip 5mu plus 5mu} \def\!{\mskip-3mu}
\def\dispmuskip{\thinmuskip= 3mu plus 0mu minus 2mu \medmuskip=  4mu plus 2mu minus 2mu \thickmuskip=5mu plus 5mu minus 2mu}
\def\textmuskip{\thinmuskip= 0mu                    \medmuskip=  1mu plus 1mu minus 1mu \thickmuskip=2mu plus 3mu minus 1mu}
\textmuskip
\def\beq{\dispmuskip\begin{equation}}    \def\eeq{\end{equation}\textmuskip}
\def\beqn{\dispmuskip\begin{displaymath}}\def\eeqn{\end{displaymath}\textmuskip}
\def\bqa{\dispmuskip\begin{eqnarray}}    \def\eqa{\end{eqnarray}\textmuskip}
\def\bqan{\dispmuskip\begin{eqnarray*}}  \def\eqan{\end{eqnarray*}\textmuskip}

\newenvironment{keywords}{\centerline{\small\bf
Keywords}\vspace{0.5ex}\begin{quote}\small}{\par\end{quote}\vskip
1ex}

\newtheorem{theorem}{Theorem}
\newtheorem{corollary}{Corollary}
\newtheorem{lemma}{Lemma}
\newtheorem{definition}{Definition}

\def\ftheorem#1#2#3{\begin{theorem}[#2]\label{#1} #3 \end{theorem} }
\def\fcorollary#1#2#3{\begin{corollary}[#2]\label{#1} #3 \end{corollary} }
\def\flemma#1#2#3{\begin{lemma}[#2]\label{#1} #3 \end{lemma} }
\def\fdefinition#1#2#3{\begin{definition}[#2]\label{#1} #3 \end{definition} }

\def\toinfty#1{\stackrel{#1\to\infty}{\longrightarrow}}
\def\qed{\sqcap\!\!\!\!\sqcup}
\def\odm{{\textstyle{1\over m}}}
\def\eps{\varepsilon}
\def\pb#1{\def\pb{\underline}\underline{#1}} % probability notation
\def\best{*}                              % or {best}
\def\xiai{{\xi}}
\def\muai{{\mu}}
\def\rhoai{{\rho}}
\def\M{{\cal M}}
\def\X{{\cal X}}
\def\Y{{\cal Y}}
\def\E{{\bf E}}

\begin{document}
%%%%%%%%%%%%%%%%%%%%%%%%%%%%%%%%%%%%%%%%%%%%%%%%%%%%%%%%%%%%%%%%%
%                      T i t l e - P a g e                      %
%%%%%%%%%%%%%%%%%%%%%%%%%%%%%%%%%%%%%%%%%%%%%%%%%%%%%%%%%%%%%%%%%

\begin{titlepage}

\title{\vspace*{-2cm}
 {\small Technical Report IDSIA-04-02 \hfil$-$\hfil
         17 April 2002 \hfil$-$\hfil
         Proceedings of COLT-2002\\[5mm]}
  \Large\sc\hrule height1pt \vskip 2mm
     Self-Optimizing and Pareto-Optimal Policies
     in General Environments based on Bayes-Mixtures
\vskip 5mm \hrule height1pt}

\author{Marcus Hutter
\\[3ex]
\small IDSIA, Galleria 2, CH-6928 Manno-Lugano, Switzerland%
\footnote{This work was supported by SNF grant 2000-61847.00 to J\"urgen
Schmidhuber.} \\
\small marcus@idsia.ch $\quad-\quad$ http://www.idsia.ch/$^{_{_\sim}}\!$marcus
}

\date{}

\maketitle              % typeset the title of the contribution

\hfill

\begin{abstract}
The problem of making sequential decisions in unknown
pro\-ba\-bi\-listic environments is studied. In cycle $t$ action
$y_t$ results in perception $x_t$ and reward $r_t$, where all
quantities in general may depend on the complete history. The
perception $x_t$ and reward $r_t$ are sampled from the (reactive)
environmental probability distribution $\mu$. This very general
setting includes, but is not limited to, (partial observable, k-th
order) Markov decision processes. Sequential decision theory tells
us how to act in order to maximize the total expected reward,
called value, if $\mu$ is known. Reinforcement learning is usually
used if $\mu$ is unknown. In the Bayesian approach one defines a
mixture distribution $\xi$ as a weighted sum of distributions
$\nu\in\M$, where $\M$ is any class of distributions including the
true environment $\mu$. We show that the Bayes-optimal policy
$p^\xi$ based on the mixture $\xi$ is self-optimizing in the sense
that the average value converges asymptotically for all $\mu\in\M$
to the optimal value achieved by the (infeasible) Bayes-optimal
policy $p^\mu$ which knows $\mu$ in advance. We show that the
necessary condition that $\M$ admits self-optimizing policies at
all, is also sufficient. No other structural assumptions are made
on $\M$. As an example application, we discuss ergodic Markov
decision processes, which allow for self-optimizing policies.
Furthermore, we show that $p^\xi$ is Pareto-optimal in the sense
that there is no other policy yielding higher or equal value in
{\em all} environments $\nu\in\M$ and a strictly higher value in
at least one.
\end{abstract}

\end{titlepage}

%%%%%%%%%%%%%%%%%%%%%%%%%%%%%%%%%%%%%%%%%%%%%%%%%%%%%%%%%%%%%%%
\section{Introduction}\label{secInt}
%%%%%%%%%%%%%%%%%%%%%%%%%%%%%%%%%%%%%%%%%%%%%%%%%%%%%%%%%%%%%%%

%------------------------------%
\paragraph{Reinforcement learning:}
%------------------------------%
There exists a well developed theory for reinforcement learning
agents in known probabilistic environments (like Blackjack) called
sequential decision theory \cite{Bellman:57,Bertsekas:95}.
The optimal agent is the one which maximizes the future expected
reward sum. This setup also includes deterministic environments
(like static mazes). Even adversarial environments (like Chess or
Backgammon) may be seen as special cases in some sense
\cite[ch.6]{Hutter:00kcunai} (the reverse is also true
\cite{Brafman:00}). Sequential decision theory deals with a wide
range of problems, and provides a general formal solution in the
sense that it is mathematically rigorous and (uniquely) specifies
the optimal solution (leaving aside computational issues).
The theory breaks down when the environment is unknown (like when
driving a car in the real world). Reinforcement learning
algorithms exist for unknown Markov decision processes ({\sc
mdp}$s$) with small state space, and for other restricted classes
\cite{Kaelbling:96,Sutton:98,Bertsekas:95,Kumar:86}, but even in
these cases their learning rate is usually far from optimum.

%------------------------------%
\paragraph{Performance measures:}
%------------------------------%
In this work we are interested in general (probabilistic)
environmental classes $\M$. We assume $\M$ is given, and that the
true environment $\mu$ is in $\M$, but is otherwise unknown.
The expected reward sum (value) $V_\mu^p$ when following policy
$p$ is of central interest. We are interested in policies $\tilde
p$ which perform well (have high value) independent of what the
true environment $\mu\in\M$ is. A natural demand from an optimal
policy is that there is no other policy yielding higher or equal
value in {\em all} environments $\nu\in\M$ and a strictly higher
value in one $\nu\in\M$. We call such a property {\em
Pareto-optimality}. The other quantity of interest is how close
$V_\mu^{\tilde p}$ is to the value $V_\mu^\best$ of the optimal
(but infeasible) policy $p^\mu$ which knows $\mu$ in advance. We
call a policy whose average value converges asymptotically for all
$\mu\in\M$ to the optimal value $V_\mu^\best$ if $\mu$ is the true
environment, {\em self-optimizing}.

%------------------------------%
\paragraph{Main new results for Bayes-mixtures:}
%------------------------------%
%Specifically, we consider probabilistic environments $\nu\in\M$.
We define the Bayes-mixture $\xi$ as a weighted average of the
environments $\nu\in\M$ and analyze the properties of the
Bayes-optimal policy $p^\xi$ which maximizes the mixture value
$V_\xi$.
One can show that not all environmental classes $\M$ admit
self-optimizing policies. One way to proceed is to search for and
prove weaker properties than self-optimizingness
\cite{Hutter:00kcunai}. Here we follow a different approach:
Obviously, the least we must demand from $\M$ to have a chance of
finding a self-optimizing policy is that there exists some
self-optimizing policy $\tilde p$ at all.
%------------------------------%
%\paragraph{What's new?:}
%------------------------------%
The main new result of this work is that this necessary condition
is also sufficient for $p^\xi$ to be self-optimizing. No other
properties need to be imposed on $\M$. The other new result is
that $p^\xi$ is always Pareto-optimal, with no conditions at
all imposed on $\M$.

%------------------------------%
\paragraph{Contents:}
%------------------------------%
Section \ref{secSetup} defines the model of agents acting in
general probabilistic environments and defines the finite horizon
value of a policy and the optimal value-maximizing policy.
Furthermore, the mixture-distribution is introduced and the
fundamental linearity and convexity properties of the
mixture-values is stated.
Section \ref{secPareto} defines and proves
Pareto-optimality of $p^\xi$. The concept is refined to balanced
Pareto-optimality, showing that a small increase of the value for
some environments only leaves room for a small decrease in others.
Section \ref{secSelfOptA} shows that $p^\xi$ is self-optimizing if
$\M$ admits self-optimizing policies, and also gives the speed of
convergence in the case of finite $\M$.
The finite horizon model has several disadvantages. For this
reason Section \ref{secDiscount} defines the discounted (infinite
horizon) future value function, and the corresponding optimal
value-maximizing policy. Pareto-optimality and
self-optimizingness of $p^\xi$ are shown shown for this model.
As an application we show in Section \ref{secMDP} that the class
of ergodic {\sc mdp}$s$ admits self-optimizing policies w.r.t.\ the
undiscounted model and w.r.t.\ the discounted model if the
effective horizon tends to infinity. Together with the results
from the previous sections this shows that $p^\xi$ is
self-optimizing for erdodic {\sc mdp}$s$.
Conclusions and outlook can be found in Section \ref{secConc}.

%%%%%%%%%%%%%%%%%%%%%%%%%%%%%%%%%%%%%%%%%%%%%%%%%%%%%%%%%%%%%%%
\section{Rational Agents in Probabilistic Environments}\label{secSetup}
%%%%%%%%%%%%%%%%%%%%%%%%%%%%%%%%%%%%%%%%%%%%%%%%%%%%%%%%%%%%%%%

%------------------------------%
\paragraph{The agent model:}
%------------------------------%
A very general framework for intelligent systems is that of
rational agents \cite{Russell:95}. In cycle $k$, an agent performs
{\em action} $y_k\!\in\!\Y$ (output) which results in a {\em
perception} or {\em observation} $x_k\!\in\!\X$ (input), followed
by cycle $k\!+\!1$ and so on. We assume that the action and
perception spaces $\X$ and $\Y$ are finite. We write
$p(x_{<k})\!=\!y_{1:k}$ to denote the output
$y_{1:k}\!\equiv\!y_1...y_k$ of the agents policy $p$ on input
$x_{<k}\!\equiv\!x_1...x_{k-1}$ and similarly
$q(y_{1:k})\!=\!x_{1:k}$ for the environment $q$ in the case of
deterministic environments. We call policy $p$ and environment $q$
behaving in this way {\it chronological}. Note that policy and
environment are allowed to depend on the complete history. We do
not make any {\sc mdp} or {\sc pomdp} assumption here, and we
don't talk about states of the environment, only about
observations.
In the more general case of a {\em probabilistic environment},
given the history $y\!x_{<k}y_k\! \equiv
\!y\!x_1...y\!x_{k-1}y_k\! \equiv \!y_1x_1...y_{k-1}x_{k-1}y_k$,
the probability that the environment leads to perception $x_k$ in
cycle $k$ is (by definition) $\rho(y\!x_{<k}y\!\pb x_k)$. The
underlined argument $\pb x_k$ in $\rho$ is a random variable and
the other non-underlined arguments $y\!x_{<k}y_k$ represent
conditions.\footnote{The standard notation
$\rho(x_k|y\!x_{<k}y_k)$ for conditional probabilities destroys
the chronological order and would become quite confusing in later
expressions.} We call probability distributions like $\rhoai$ {\it
chronological}.
Since value optimizing policies can always be chosen
deterministic, there is no real need to generalize the setting to
probabilistic policies. Arbitrarily we formalize Sections
\ref{secPareto} and \ref{secSelfOptA} in terms of deterministic
policies and Section \ref{secDiscount} in terms of probabilistic
policies.

%------------------------------%
\paragraph{Value functions and optimal policies:}
%------------------------------%
The goal of the agent is to maximize future {\em rewards}, which are
provided by the environment through the inputs $x_k$. The inputs
$x_k\!\equiv\!x'_kr_k$ are divided into a regular part $x'_k$ and
some (possibly empty or delayed) reward $r_k\in[0\,,\,r_{max}]$.%
\footnote{In the reinforcement learning literature when dealing with
{\sc (po)mdp}$s$ the reward is usually considered to be a function of the
environmental state. The zero-assumption analogue here is that the
reward $r_k$ is some probabilistic function $\rho'$ depending on
the complete history. It is very convenient to integrate $r_k$
into $x_k$ and $\rho'$ into $\rho$.}
We use the abbreviation
\bqa\label{bayes2}
  \rho(y\!x_{<k}y\!\pb x_{k:m}) \;=\;
  \rho(y\!x_{<k}y\!\pb x_k)\!\cdot\!\rho(y\!x_{1:k} y\!\pb x_{k+1})\!\cdot...\cdot\!
  \rho(y\!x_{<m}y\!\pb x_m),
\eqa
which is essentially Bayes rules, and $\eps=y\!x_{<1}$ for the
empty string. The $\rho$-expected reward sum (value) of future
cycles $k$ to $m$ with outputs $y_{k:m}$ generated by the agent's
policy $p$, the optimal policy $p^\rho$ which maximizes the value,
its action $y_k$ and the corresponding value can formally be
defined as follows.

%------------------------------%
\fdefinition{defValp}{Value function and optimal policy}{
%------------------------------%
We define the {\em value} of policy $p$ in environment $\rho$
given history $y\!x_{<k}$, or shorter, the $\rho$-value of $p$ given
$y\!x_{<k}$, as
\beq\label{defrhoVal}
  \!V_{km}^{p\rho}(y\!x_{<k}) \;:=\;
  \sum_{x_{k:m}}(r_k\!+...+\!r_m)\rho(y\!x_{<k}y\!\pb
  x_{k:m})_{|y_{1:m}=p(x_{<m})}.
\eeq
$m$ is the {\em lifespan} or initial {\em horizon} of the agent.
The $\rho$-optimal policy $p^\rho$ which maximizes the (total) value
$V_\rho^p:=V_{1m}^{p\rho}(\eps)$ is
\beq\label{defAIrhoF}
  p^\rho := \arg\max_p V_\rho^p, \quad
  V_{km}^{\best\rho}(y\!x_{<k}) :=
  V_{km}^{p^\rho\rho}(y\!x_{<k}).
\eeq
Explicit expressions for the action $y_k$ in cycle $k$ of the
$\rho$-optimal policy $p^\rho$ and their value
$V_{km}^{\best\rho}(y\!x_{<k})$ are
\beq\label{defAIrhoy}
  y_k \;=\;
  \arg\max_{y_k}\sum_{x_k}\max_{y_{k+1}}\sum_{x_{k+1}}\;...\;
  \max_{y_m}\sum_{x_m}
  (r_k\!+...+\!r_m) \!\cdot\!
  \rho(y\!x_{<k}y\!\pb x_{k:m}),
\eeq
\beq\label{defAIrhoV}
  V_{km}^{\best\rho}(y\!x_{<k}) \;=\;
  \max_{y_k}\sum_{x_k}\max_{y_{k+1}}\sum_{x_{k+1}}\;...\;
  \max_{y_m}\sum_{x_m}
  (r_k\!+...+\!r_m) \!\cdot\!
  \rho(y\!x_{<k}y\!\pb x_{k:m}).
\eeq
where $y\!x_{<k}$ is the actual history.
}%------------------------------%
One can show \cite{Hutter:00kcunai} that these definitions are
consistent and correctly capture our intention. For instance,
consider the expectimax expression (\ref{defAIrhoV}): The best
expected reward is obtained by averaging over possible perceptions
$x_i$ and by maximizing over the possible actions $y_i$. This has
to be done in chronological order $y_kx_k...y_mx_m$ to correctly
incorporate the dependency of $x_i$ and $y_i$ on the history.
Obviously
\beq\label{Vdom}
  V_{km}^{\best\rho}(y\!x_{<k}) \;\geq\;
  V_{km}^{p\rho}(y\!x_{<k})\;\forall p,
  \quad\mbox{especially}\quad
  V_\rho^\best\;\geq\;V_\rho^p\;\forall p.
\eeq

%------------------------------%
\paragraph{Known environment $\mu$:}
%------------------------------%
Let us now make a change in conventions and assume that $\mu$ is
the true environment in which the agent operates and that we know
$\mu$ (like in Blackjack).\footnote{If the existence of true
objective probabilities violates the philosophical attitude of the
reader he may assume a deterministic environment $\mu$.} Then,
policy $p^\mu$ is optimal in the sense that no other policy for an
agent leads to higher $\muai$-expected reward.
This setting includes as special cases deterministic environments,
Markov decision processes ({\sc mdp}$s$), and even adversarial
environments for special choices of $\mu$ \cite{Hutter:00kcunai}.
There is no principle problem in determining the optimal action
$y_k$ as long as $\muai$ is known and computable and $\X$, $\Y$
and $m$ are finite.

%------------------------------%
\paragraph{The mixture distribution $\xi$:}
%------------------------------%
Things drastically change if $\muai$ is unknown. For
(parameterized) {\sc mdp}$s$ with small state (parameter) space,
suboptimal reinforcement learning algorithms may be used to learn
the unknown $\muai$
\cite{Kaelbling:96,Sutton:98,Bertsekas:95,Kumar:86}. In the
Bayesian approach the true probability distribution $\muai$ is not
learned directly, but is replaced by a Bayes-mixture $\xiai$. Let
us assume that we know that the true environment $\mu$ is
contained in some known set $\M$ of environments. For convenience
we assume that $\M$ is finite or countable. The Bayes-mixture
$\xi$ is defined as
\beq\label{xiaidef}
  \xi(y\!\pb x_{1:m})= \sum_{\nu\in\M}w_\nu \nu(y\!\pb x_{1:m})
  \quad\mbox{with}\quad
  \sum_{\nu\in\M}w_\nu=1, \quad
  w_\nu>0\quad\forall\nu\in\M
\eeq
The weights $w_\nu$ may be interpreted as the prior degree of
belief that the true environment is $\nu$. Then $\xi(y\!\pb
x_{1:m})$ could be interpreted as the prior subjective belief
probability in observing $x_{1:m}$, given actions $y_{1:m}$. It
is, hence, natural to follow the policy $p^\xi$ which maximizes
$V_\xi^p$. If $\mu$ is the true environment the expected reward
when following policy $p^\xi$ will be $V_\mu^{p^\xi}$. The optimal
(but infeasible) policy $p^\mu$ yields reward $V_\mu^{p^\mu}\equiv
V_\mu^{\best}$. It is now of interest $(a)$ whether there are
policies with uniformly larger value than $V_\mu^{p^\xi}$ and
$(b)$ how close $V_\mu^{p^\xi}$ is to $V_\mu^{\best}$. These are
the main issues of the remainder of this work.
%The side remark in the following paragraph may be skipped.

%------------------------------%
\paragraph{A universal choice of $\xi$ and $\M$:}
%------------------------------%
One may also ask what the most general class $\M$ and weights
$w_\nu$ could be. Without any prior knowledge we should include
{\em all} environments in $\M$. In this generality this approach
leads at best to negative results. More useful is the assumption
that the environment possesses some structure, we just don't know
which. From a computational point of view we can only unravel
effective structures which are describable by (semi)computable
probability distributions. So we may include {\em all}
(semi)computable (semi)distribu\-tions in $\M$. Occam's razor
tells us to assign high prior belief to simple environments. Using
Kolmogorov's universal complexity measure $K(\nu)$ for
environments $\nu$ one should set $w_\nu\sim 2^{-K(\nu)}$, where
$K(\nu)$ is the length of the shortest program on a universal
Turing machine computing $\nu$. The resulting policy $p^\xi$ has
been developed and intensively discussed in
\cite{Hutter:00kcunai}. It is a unification of sequential decision
theory \cite{Bellman:57,Bertsekas:95} and Solomonoff's celebrated
universal induction scheme \cite{Solomonoff:78,Li:97}. In the
following we consider generic $\M$ and $w_\nu$.
The following property of $V_\rho$ is crucial.

%------------------------------%
\ftheorem{thVlinconv}{Linearity and convexity of $V_\rho$ in $\rho$}{
%------------------------------%
$V_\rho^p$ is a linear function in $\rho$ and $V_\rho^\best$
is a convex function in $\rho$ in the sense that
\beqn%\label{thV1lin}\label{thV1conv}
  V_\xi^p \;= \sum_{\nu\in\M}w_\nu V_\nu^p
  \quad\mbox{and}\quad
  V_\xi^\best \;\leq \sum_{\nu\in\M}w_\nu V_\nu^\best
  \quad\mbox{where}\quad
  \xi(y\!\pb x_{1:m})= \sum_{\nu\in\M}w_\nu \nu(y\!\pb x_{1:m})
\eeqn\vspace{-1ex}
}%------------------------------%

\noindent{\bf Proof:} Linearity is obvious from the definition
of $V_\rho^p$. Convexity follows from
$V_\xi^\best\equiv V_\xi^{p^\xi}=\sum_\nu w_\nu V_\nu^{p^\xi}\leq
\sum_\nu w_\nu V_\nu^\best$, where the identity is definition
(\ref{defAIrhoF}), the equality uses linearity of $V_\rho^{p^\xi}$
just proven, and the last inequality follows from the dominance
(\ref{Vdom}) and non-negativity of the weights $w_\nu$. $\qed$

One loose interpretation of the convexity is that a mixture can
never increase performance. In the remainder of this work $\mu$
denotes the true environment, $\rho$ any distribution, and $\xi$
the Bayes-mixture of distributions $\nu\in\M$.

%%%%%%%%%%%%%%%%%%%%%%%%%%%%%%%%%%%%%%%%%%%%%%%%%%%%%%%%%%%%%%%
\section{Pareto Optimality of policy $p^\xi$}\label{secPareto}
%%%%%%%%%%%%%%%%%%%%%%%%%%%%%%%%%%%%%%%%%%%%%%%%%%%%%%%%%%%%%%%

The total $\mu$-expected reward $V_\mu^{p^\xi}$ of policy $p^\xi$
is of central interest in judging the
performance of policy $p^\xi$. We know that there {\em are} policies
(e.g.\ $p^\mu$) with higher $\mu$-value
($V_\mu^\best\geq V_\mu^{p^\xi}$). In general, every policy based
on an estimate $\rho$ of $\mu$ which is closer to $\mu$ than $\xi$
is, outperforms $p^\xi$ in environment $\mu$, simply because it is
more taylored toward $\mu$. On the other hand, such a system
probably performs worse than $p^\xi$ in other environments.
Since we do not know $\mu$ in advance we may ask
whether there exists a policy $p$ with better or equal performance
than $p^\xi$ in {\em all} environments $\nu \in\M$ and a strictly
better performance for one $\nu \in\M$. This would clearly
render $p^\xi$ suboptimal. We show that there is no such $p$.

%------------------------------%
\ftheorem{thaiPareto}{Pareto optimality}{
%------------------------------%
Policy $p^\xi$ is Pareto-optimal in the sense that there is no
other policy $p$ with $V_{\nu}^p\geq V_{\nu}^{p^\xi}$ for all $\nu
\in \M$ and strict inequality for at least one $\nu$.
}%------------------------------%

\noindent{\bf Proof:} We want to arrive at a contradiction by
assuming that $p^\xi$ is not Pareto-optimal, i.e.\ by assuming the
existence of a policy $p$ with $V_\nu^p\geq V_\nu^{p^\xi}$ for all
$\nu \in \M$ and strict inequality for at least one $\nu$:
\beqn
  V_\xi^p \;=\;
  \sum_\nu w_\nu V_\nu^p \;>\;
  \sum_\nu w_\nu V_\nu^{p^\xi} \;=\;
  V_\xi^{p^\xi} \;\equiv\;
  V_\xi^\best \;\geq\;
  V_\xi^p
\eeqn
The two equalities follow from linearity of $V_\rho$ (Theorem
\ref{thVlinconv}). The strict inequality follows from the
assumption and from $w_\nu>0$. The identity is just Definition
\ref{defValp}(\ref{defAIrhoF}). The last inequality follows from
the fact that $p^\xi$ maximizes by definition the universal value
(\ref{Vdom}). The contradiction $V_\xi^p > V_\xi^p$ proves
Pareto-optimality of policy $p^\xi$.$\qed$

Pareto-optimality should be regarded as a necessary condition for
an agent aiming to be optimal. From a practical point
of view a significant increase of $V$ for many environments
$\nu$ may be desirable even if this causes a small decrease of
$V$ for a few other $\nu$. The impossibility of such a
``balanced'' improvement is a more demanding condition on $p^\xi$
than pure Pareto-optimality. The next theorem
shows that $p^\xi$ is also balanced-Pareto-optimal in
the following sense:

%------------------------------%
\ftheorem{thaiElPareto}{Balanced Pareto optimality}{
%------------------------------%
\beqn
  \Delta_\nu:=V_\nu^{p^\xi}-V_\nu^{\tilde p} ,\quad
  \Delta:=\sum_{\nu\in\M}w_\nu\Delta_\nu \quad\Rightarrow\quad
  \Delta\geq 0.
\eeqn
This implies the following: Assume $\tilde p$ has lower value than
$p^\xi$ on environments $\cal L$ by a
total weighted amount of $\Delta_{\cal
L} := \sum_{\lambda\in\cal L}w_\lambda\Delta_\lambda$. Then
$\tilde p$ can have higher value on $\eta \in {\cal
H} := \M \setminus \cal L$, but the improvement is bounded by
$\Delta_{\cal H} := |\sum_{\eta\in\cal
H}w_\eta\Delta_\eta| \leq \Delta_{\cal L}$. Especially
$|\Delta_\eta| \leq w_\eta^{-1}\max_{\lambda\in\cal
L}\Delta_\lambda$.
}%------------------------------%
This means that a weighted value increase $\Delta_{\cal H}$ by
using $\tilde p$ instead of $p^\xi$ is compensated by an at least
as large weighted decrease $\Delta_{\cal L}$ on other
environments. If the decrease is small, the increase can also only
be small. In the special case of only a single environment with
decreased value $\Delta_\lambda$, the increase is bound by
$\Delta_\eta \leq {w_\lambda\over w_\eta}|\Delta_\lambda|$, i.e.\
a decrease by an amount $\Delta_\lambda$ can only cause an
increase by at most the same amount times a factor
${w_\lambda\over w_\eta}$. For the choice of the weights
$w_\nu\sim 2^{-K(\nu)}$, a decrease can only cause a smaller
increase in simpler environments, but a scaled increase in more
complex environments. Finally note that pure Pareto-optimality
(Theorem \ref{thaiPareto}) follows from balanced Pareto-optimality
in the special case of no decrease $\Delta_{\cal L} \equiv 0$.

\noindent{\bf Proof:} $\Delta \geq 0$ follows from
$\Delta=\sum_\nu w_\nu[V_\nu^{p^\xi} - V_\nu^{\tilde
p}]=V_\xi^{p^\xi} - V_\xi^{\tilde p}\geq 0$, where we have used
linearity of $V_\rho$ (Theorem \ref{thVlinconv}) and dominance
$V_\xi^{p^\xi} \geq V_\xi^p$ (\ref{Vdom}). The remainder of
Theorem \ref{thaiElPareto} is obvious from
$0 \leq \Delta = \Delta_{\cal L} - \Delta_{\cal H}$ and by
bounding the weighted average $\Delta_\eta$ by its maximum.$\qed$

%%%%%%%%%%%%%%%%%%%%%%%%%%%%%%%%%%%%%%%%%%%%%%%%%%%%%%%%%%%%%%%
\section{Self-optimizing Policy $p^\xi$ w.r.t.\ Average Value}\label{secSelfOptA}
%%%%%%%%%%%%%%%%%%%%%%%%%%%%%%%%%%%%%%%%%%%%%%%%%%%%%%%%%%%%%%%

In the following we study under which circumstances\footnote{%
Here and elsewhere we interpret $a_m\to b_m$ as an abbreviation
for $a_m-b_m\to 0$. $\lim_{m\to\infty}b_m$ may not exist.}
\beq\label{VxitoV}
  \odm V_{1m}^{p^\xi\nu}\to\odm V_{1m}^{\best\nu}
  \quad\mbox{for}\quad m\to\infty
  \quad\mbox{for {\em all}}\quad \nu\in\M.
\eeq
The least we must demand from $\M$ to have a chance that
(\ref{VxitoV}) is true is that there exists some policy $\tilde p$ at all with this
property, i.e.\
\beq\label{VptoV}
  \exists\tilde p:\;\odm V_{1m}^{\tilde p\nu}\to\odm V_{1m}^{\best\nu}
  \quad\mbox{for}\quad m\to\infty
  \quad\mbox{for {\em all}}\quad \nu\in\M.
\eeq
Luckily, this necessary condition will also be sufficient. This is
another (asymptotic) optimality property of policy $p^\xi$. {\em
If} universal convergence in the sense of (\ref{VptoV}) is
possible at all in a class of environments $\M$, then policy
$p^\xi$ converges in the sense of (\ref{VxitoV}). We will call
policies $\tilde p$ with a property like (\ref{VptoV}) {\em
self-optimizing} \cite{Kumar:86}.
The following two Lemmas pave the way for proving the convergence
Theorem.

%------------------------------%
\flemma{lemVDconv}{Value difference relation}{
%------------------------------%
\beqn
  0 \;\leq\; V_\nu^\best-V_\nu^{\tilde p} \;=:\; \Delta_\nu
  \quad\Rightarrow\quad
  0 \;\leq\; V_\nu^\best-V_\nu^{p^\xi} \;\leq\; {\textstyle{1\over w_\nu}}\Delta
  \quad\mbox{with}\quad
  \Delta:=\sum_{\nu \in\M}w_\nu\Delta_\nu
\eeqn\vspace{-3ex}
}%------------------------------%

\noindent{\bf Proof:} The following sequence of inequalities proves the
lemma:
\beqn\textstyle
  0 \;\leq\; w_\nu[V_\nu^\best\!-\!V_\nu^{p^\xi}] \;\leq\;
%  \sum_\nu \underbrace{w_\nu}_{\geq 0}
%  [\underbrace{V_\nu^\best\!-\!V_\nu^{p^\xi}}_{\geq 0}] \;\leq\;
  \sum_\nu w_\nu [V_\nu^\best\!-\!V_\nu^{p^\xi}] \;\leq\;
  \sum_\nu w_\nu[V_\nu^\best\!-\!V_\nu^{\tilde p}] \;=\;
  \sum_\nu w_\nu\Delta_\nu \;\equiv\; \Delta
\eeqn
In the first and second inequality we used $w_\nu \geq 0$ and
$V_\nu^\best - V_\nu^{p^\xi} \geq 0$. The last inequality
follows from
$
  \sum_\nu w_\nu V_\nu^{p^\xi}  =
  V_\xi^{p^\xi}  \equiv   V_\xi^\best  \geq  V_\xi^{\tilde p}  =
  \sum_\nu w_\nu V_\nu^{\tilde p}
$. $\qed$

We also need some results for averages of functions
$\delta_\nu(m)\geq 0$ converging to zero.

%------------------------------%
\flemma{lemVdconv}{Convergence of averages}{
%------------------------------%
For $\delta(m):=\sum_{\nu\in\M}w_\nu\delta_\nu(m)$ the following
holds (we only need $\sum_\nu w_\nu\leq 1$):
\beqn
\begin{array}{rllllll}
    i)  & \displaystyle
          \delta_\nu(m)\leq f(m)
        & \forall\nu
        & \mbox{implies}
        & \delta(m) \leq f(m). \\[1ex]
   ii)  & \displaystyle
          \delta_\nu(m)\toinfty{m} 0
        & \forall\nu
        & \mbox{implies}
        & \delta(m)\toinfty{m} 0
        & \mbox{if}
        & 0\leq\delta_\nu(m)\leq c. \\[1ex]
%  iii)  & \displaystyle
%          \delta(m)\leq\max_\nu\delta_\nu(m).
\end{array}
\eeqn\vspace{-3ex}
}%------------------------------%

\noindent{\bf Proof:} $(i)$ immediately follows from
$\delta(m)=\sum_\nu w_\nu\delta_\nu(m) \leq
\sum_\nu w_\nu f(m) \leq f(m)$.
For $(ii)$ we choose some order on $\M$ and some $\nu_0 \in \M$
large enough such that $\sum_{\nu\geq\nu_0}w_\nu\leq{\eps\over c}$.
Using $\delta_\nu(m) \leq c$ this implies
\beqn
  \sum_{\nu\geq\nu_0}w_\nu\delta_\nu(m) \;\leq\;
  \sum_{\nu\geq\nu_0}w_\nu c \;\leq\; \eps.
\eeqn
Furthermore, the assumption $\delta_\nu(m) \to 0$ means
that there is an $m_{\nu\eps}$ depending on $\nu$ and $\eps$ such
that $\delta_\nu(m)\leq\eps$ for all $m \geq m_{\nu\eps}$. This
implies
\beqn
  \sum_{\nu\leq\nu_0}w_\nu\delta_\nu(m) \;\leq\;
  \sum_{\nu\leq\nu_0}w_\nu \eps \;\leq\; \eps
  \quad\mbox{for all}\quad
  m \;\geq\; \max_{\nu\leq\nu_0}\{m_{\nu\eps}\} \;=\; :m_\eps.
  \vspace{-1ex}
\eeqn
$m_\eps < \infty$, since the maximum is over a finite set.
Together we have
\beqn
  \delta(m) \;\equiv\;
  \sum_{\nu\!\in\!\M}w_\nu\delta_\nu(m) \;\leq\; 2\eps
  \quad\mbox{for}\quad
   m \;\geq\; m_\eps
  \quad\Rightarrow\quad
  \delta(m)\to 0
  \quad\mbox{for}\quad
  m\to\infty\vspace{-1ex}
\eeqn
since $\eps$ was arbitrary and $\delta(m) \geq 0$. $\qed$

%------------------------------%
\ftheorem{thVcconv}{Self-optimizing policy $p^\xi$ w.r.t.\ average value}{
%------------------------------%
There exists a sequence of policies $\tilde p_m$, $m=1,2,3,...$
with value within $\Delta(m)$ to optimum for all environments
$\nu\in\M$, then, save for a constant factor, this also holds for
the sequence of universal policies $p^\xi_m$,  i.e.\
\beqn
i)\quad\mbox{If}\quad
 \exists\tilde p_m\forall\nu:
 V_{1m}^{\best\nu}-V_{1m}^{\tilde p_m\nu} \leq \Delta(m)
 \quad\Longrightarrow\quad
 V_{1m}^{\best\mu}-V_{1m}^{p^\xi_m\mu} \leq
 \textstyle{1\over w_\mu}\Delta(m).
\eeqn
If there exists a sequence of self-optimizing
policies $\tilde p_m$ in the sense that their expected average
reward $\odm V_{1m}^{\tilde p_m\nu}$ converges to the optimal
average $\odm V_{1m}^{\best\nu}$ for all environments $\nu\in\M$,
then this also holds for the sequence of universal policies
$p^\xi_m$, i.e.\
\beqn
ii) \quad\mbox{If}\quad
  \exists\tilde p_m\forall\nu:
 \odm V_{1m}^{\tilde p_m\nu} \;\toinfty{m}\; \odm V_{1m}^{\best\nu}
 \quad\Longrightarrow\quad
 \odm V_{1m}^{p^\xi_m\mu} \;\toinfty{m}\; \odm V_{1m}^{\best\mu}.
 \rule{2ex}{0ex}
\eeqn\vspace{-3ex}
}%------------------------------%

The beauty of this theorem is that if universal convergence in the
sense of (\ref{VptoV}) is possible at all in a class of
environments $\M$, then policy $p^\xi$ converges (in the sense of
(\ref{VxitoV})). The necessary condition of convergence is also
sufficient. The unattractive point is that this is not an
asymptotic convergence statement for $V_{km}^{p^\xi \mu}$ of a
single policy $p^\xi$ for $k\to\infty$ for some fixed $m$, and in
fact no such theorem could be true, since always $k\leq m$. The
theorem merely says that under the stated conditions the average
value of $p^\xi_m$ can be arbitrarily close to optimum for
sufficiently large (pre-chosen) horizon $m$. This weakness will be
resolved in the next subsection.

\noindent{\bf Proof:} $(i)$ $\Delta_\nu(m)=f(m)$ implies
$\Delta(m)=f(m)$ by Lemma \ref{lemVdconv}$(i)$.
Inserting this in Lemma \ref{lemVDconv} proves Theorem
\ref{thVcconv}$(i)$ (recovering the $m$ dependence and finally
renaming $f \leadsto \Delta$).

$(ii)$ We define $\delta_\nu(m) := \odm\Delta_\nu(m) = \odm
[V_\nu^\best-V_\nu^{\tilde p}]$. Since we assumed bounded rewards
$0 \leq r \leq r_{max}$ we have
\beqn
  V_\nu^* \leq m r_{max}
  \quad\mbox{and}\quad
  V_\nu^{\tilde p} \geq 0
  \quad\Rightarrow\quad
  \Delta_\nu \leq m r_{max}
  \quad\Rightarrow\quad
  0 \leq \delta_\nu(m) \leq c:=r_{max}.
\eeqn
The premise in Theorem \ref{thVcconv}$(ii)$ is that
$
  \delta_\nu(m)=\odm[V_{1m}^{\best\nu} - V_{1m}^{\tilde
  p\nu}]\to 0
$
which implies
\beqn\textstyle
  0 \;\leq\; \odm[V_{1m}^{\best\nu}-V_{1m}^{p^\xi\nu}]
  \;\leq\;
  {1\over w_\nu}{\Delta(m)\over m} \;=\; {1\over w_\nu}\delta(m)
  \;\to\; 0.
\eeqn
The inequalities follow from Lemma \ref{lemVDconv} and convergence
to zero from Lemma \ref{lemVdconv}$(ii)$.
This proves Theorem \ref{thVcconv}$(ii)$. $\qed$.

In Section \ref{secMDP} we show that a converging $\tilde p$
exists for ergodic {\sc mdp}$s$, and hence $p^\xi$ converges in
this environmental class too (in the sense of Theorem
\ref{thVcconv}).

%%%%%%%%%%%%%%%%%%%%%%%%%%%%%%%%%%%%%%%%%%%%%%%%%%%%%%%%%%%%%%%
\section{Discounted Future Value Function}\label{secDiscount}
%%%%%%%%%%%%%%%%%%%%%%%%%%%%%%%%%%%%%%%%%%%%%%%%%%%%%%%%%%%%%%%

We now shift our focus from the total value $V_{1m}$, $m\to\infty$
to the future value (value-to-go) $V_{k?}$, $k\to\infty$. The main
reason is that we want to get rid of the horizon parameter $m$. In
the last subsection we have shown a convergence theorem for
$m\to\infty$, but a specific policy $p^\xi$ is defined for all
times relative to a fixed horizon $m$. Current time $k$ is moving,
but $m$ is fixed\footnote{A dynamic horizon like $m\leadsto
m_k=k^2$ can lead to policies with very poor performance
\cite[Ch.4]{Hutter:00kcunai}.}. Actually, to use $k\to\infty$
arguments we {\em have} to get rid of $m$, since $k\leq m$. This
is the reason for the question mark in $V_{k?}$ above.

We eliminate the horizon by discounting the rewards
$r_k\leadsto\gamma_k r_k$ with $\sum_{i=1}^\infty\gamma_i<\infty$
and letting $m\to\infty$. The analogue of $m$ is now an effective
horizon $h_k^{e\!f\!f}$ which may be defined by
$\sum_{i=k}^{k+h_k^{e\!f\!f}}\!\!\gamma_k \sim
\sum_{i=k+h_k^{e\!f\!f}}^\infty\; \gamma_k$.
See \cite[Ch.4]{Hutter:00kcunai} for a detailed discussion of the
horizon problem. Furthermore, we renormalize $V_{k\infty}$ by
$\sum_{i=k}^\infty\gamma_i$ and denote it by $V_{k\gamma}$. It can
be interpreted as a future expected weighted-average reward.
Furthermore we extend the definition to probabilistic policies
$\pi$.

%------------------------------%
\fdefinition{defAIrhoDisc}{Discounted value function and optimal policy}{
%------------------------------%
We define the $\gamma$ discounted weighted-average future {\em
value} of (probabilistic) policy $\pi$ in environment $\rho$ given
history $y\!x_{<k}$, or shorter, the $\rho$-value of $\pi$ given
$y\!x_{<k}$, as
\beqn
  V_{k\gamma}^{\pi\rho}(y\!x_{<k}) \;:=\;
  {1\over\Gamma_k}\lim_{m\to\infty}
  \sum_{y\!x_{k:m}}(\gamma_k r_k\!+...+\!\gamma_m r_m)
  \rho(y\!x_{<k}y\!\pb x_{k:m}) \pi(y\!x_{<k}\pb y\!x_{k:m})
\eeqn
with $\Gamma_k:=\sum_{i=k}^\infty\gamma_i$. The policy $p^\rho$ is defined as to
maximize the future value $V_{k\gamma}^{\pi\rho}$:
\beqn
  p^\rho := \arg\max_\pi V_{k\gamma}^{\pi\rho}, \qquad
  V_{k\gamma}^{\best\rho} :=
  V_{k\gamma}^{p^\rho\rho} =
  \max_\pi V_{k\gamma}^{\pi\rho} \geq
  V_{k\gamma}^{\pi\rho}\,\forall\pi.
\eeqn\vspace{-3ex}
}%------------------------------%

\pagebreak[2]
\noindent{\bf Remarks:}\vspace{-2ex}
\begin{itemize}\parskip=0ex\parsep=0ex\itemsep=0ex
\item $\pi(y\!x_{<k}\pb y\!x_{k:m})$ is actually independent of $x_m$, since $\pi$ is
chronological.
\item Normalization of $V_{k\gamma}$ by $\Gamma_k$ does not affect the policy $p^\rho$.
\item The definition of $p^\rho$ is independent of $k$.%
\item Without normalization by $\Gamma_k$ the future values would converge to
zero for $k\to\infty$ in every environment for every policy.
\item For an {\sc mdp} environment, a stationary policy, and geometric
discounting $\gamma_k\sim\gamma^k$, the future value is
independent of $k$ and reduces to the well-known {\sc mdp} value
function.
\item There is always a
deterministic optimizing policy $p^\rho$ (which we use).
\item For a deterministic policy there is exactly one $y_{k:m}$
for each $x_{k:m}$ with $\pi\neq 0$. The sum over $y_{k:m}$ drops
in this case.
\item An iterative representation as in Definition \ref{defValp} is possible.
\item Setting $\gamma_k=1$ for $k\leq m$ and $\gamma_k=0$ for $k>m$
gives back the undiscounted model (\ref{defValp}) with
$V_{1\gamma}^{p\rho}={1\over m}V_{1m}^{p\rho}$.
\item $V_{k\gamma}$ (and $w_k^\nu$ defined below) depend on the realized
history $y\!x_{<k}$.
\end{itemize}
Similarly to the previous sections one can prove the following
properties:

%------------------------------%
\ftheorem{thVlinconv2}{Linearity and convexity of $V_\rho$ in $\rho$}{
%------------------------------%
$V_{k\gamma}^{\pi\rho}$ is a linear function in $\rho$ and
$V_{k\gamma}^{\best\rho}$ is a convex function in $\rho$ in the
sense that
\beqn
  V_{k\gamma}^{\pi\xi} \;= \sum_{\nu\in\M}w_k^\nu\,V_{k\gamma}^{\pi\nu}
  \quad\mbox{and}\quad
  V_{k\gamma}^{\best\xi} \;\leq \sum_{\nu\in\M}w_k^\nu\,V_{k\gamma}^{\best\nu}
\eeqn\vspace{-2ex}
\beqn
  \mbox{where}\quad
  \xi(y\!x_{<k}y\!\pb x_{k:m})= \sum_{\nu\in\M}w_k^\nu\,\nu(y\!x_{<k}y\!\pb
  x_{k:m})
  \quad\mbox{with}\quad
  w_k^\nu:=w_\nu{\nu(y\!\pb x_{<k})\over\xi(y\!\pb x_{<k})}
\eeqn
\vspace{-2ex}
}%------------------------------%
The conditional representation of $\xi$ can be proven by dividing
the definition (\ref{xiaidef}) of $\xi(y\!\pb x_{1:m})$ by
$\xi(y\!\pb x_{<k})$ and by using Bayes rules (\ref{bayes2}). The
posterior weight $w_k^\nu$ may be interpreted as the posterior
belief in $\nu$ and is related to learning aspects of policy
$p^\xi$.

%------------------------------%
\ftheorem{thaiPareto2}{Pareto optimality}{
%------------------------------%
For every $k$ and history $y\!x_{<k}$ the following holds:
$p^\xi$ is Pareto-optimal in the sense that there is no other
policy $\pi$ with $V_{k\gamma}^{\pi\nu}\geq
V_{k\gamma}^{p^\xi\nu}$ for all $\nu\in\M$ and strict
inequality for at least one $\nu$.
}%------------------------------%

%------------------------------%
\flemma{lemVDconv2}{Value difference relation}{
%------------------------------%
\beqn
  0 \leq V_{k\gamma}^{\best\nu}-V_{k\gamma}^{\tilde \pi_k\nu}
  =: \Delta_k^\nu
  \quad\Rightarrow\quad
  0 \leq V_{k\gamma}^{\best\nu}-V_{k\gamma}^{p^\xi\nu}
  \leq {\textstyle{1\over w_k^\nu}}\Delta_k
  \quad\mbox{with}\quad
  \Delta_k:=\sum_{\nu\in\M}w_k^\nu\Delta_k^\nu
\eeqn\vspace{-2ex}
}%------------------------------%
The proof of Theorem \ref{thaiPareto2} and Lemma \ref{lemVDconv2}
follows the same steps as for Theorem \ref{thaiPareto} and Lemma \ref{lemVDconv}
with appropriate replacements. The proof of the analogue of the
convergence Theorem \ref{thVcconv} involves one additional step.
We abbreviate ``with $\mu$ probability 1'' by w.$\mu$.p.1.

%------------------------------%
\ftheorem{thVcconv2}{Self-optimizing policy $p^\xi$ w.r.t.\ discounted value}{
%------------------------------%
For any $\M$, if there exists a sequence of self-optimizing
policies $\tilde \pi_k$ $k=1,2,3,...$ in the sense that their
expected weighted-average reward $V_{k\gamma}^{\tilde
\pi_k\nu}$ converges for $k\to\infty$ with
$\nu$-probability one to the optimal value
$V_{k\gamma}^{\best\nu}$ for all environments
$\nu\in\M$, then this also holds for the universal policy $p^\xi$
in the true $\mu$-environment, i.e.\
\beqn
 \quad \mbox{If }
  \exists\tilde \pi_k\forall\nu:
 V_{k\gamma}^{\tilde \pi_k\nu} \;\toinfty{k}\; V_{k\gamma}^{\best\nu}
 \quad\mbox{w.$\nu$.p.1} \quad\Longrightarrow\quad
 V_{k\gamma}^{p^\xi\mu} \;\toinfty{k}\; V_{k\gamma}^{\best\mu}
 \quad\mbox{w.$\mu$.p.1}.
\eeqn
The probability qualifier refers to the historic perceptions
$x_{<k}$. The historic actions $y_{<k}$ are arbitrary.
}%------------------------------%
The conclusion is valid for action histories $y_{<k}$ if the
condition is satisfied for this action history. Since we usually
need the conclusion for the $p^\xi$-action history, which is hard
to characterize, we usually need to prove the condition for {\em
all} action histories.
Theorem \ref{thVcconv2} is a powerful result: An (inconsistent)
sequence of probabilistic policies $\tilde\pi_k$ suffices to prove
the existence of a (consistent) deterministic policy $p^\xi$.
A result similar to Theorem \ref{thVcconv}$(i)$ also holds for the
discounted case, roughly saying that
$V^{\tilde\pi}-V^\best=O(\Delta(k))$ implies
$V^{p^\xi}-V^\best={1\over\eps}O(\Delta(k))$ with $\mu$ probability
$1-\eps$ for finite $\M$.

\noindent{\bf Proof:}
We define $\delta_\nu(k) := \Delta_k^\nu =
V_{k\gamma}^{\best\nu}-V_{k\gamma}^{\tilde\pi\nu}$. Since we
assumed bounded rewards $0 \leq r \leq r_{max}$ and $V_{k\gamma}^{\best\nu}$ is a weighted
average of rewards we have
\beqn
  V_{k\gamma}^{\best\mu} \leq r_{max}
  \quad\mbox{and}\quad
  V_{k\gamma}^{\tilde\pi\mu} \geq 0
  \quad\Rightarrow\quad
  0 \leq \delta_\nu(k) = \Delta_k^\nu \leq c:=r_{max}.
\eeqn
The following inequalities follow from Lemma \ref{lemVDconv2}:
\beq\label{dvpr}\textstyle
  0 \;\leq\; V_{k\gamma}^{\best\mu} - V_{k\gamma}^{p^\xi\mu}
  \;\leq\;
  {1\over w_k^\mu} \Delta_k \;=\; {1\over w_k^\mu}\delta(k)
  \;\stackrel?\to 0
\eeq
The premise in Theorem \ref{thVcconv2} is that
$
  \delta_\nu(k)=V_{k\gamma}^{\best\nu} - V_{k\gamma}^{\tilde \pi\nu} \to 0
$
for $k\to\infty$ which implies $\delta(k)\to 0$ (w.$\mu$.p.1) by
Lemma \ref{lemVdconv}$(ii)$. What is new and what remains to be
shown is that $w_k^\mu$ is bounded from below in order
to have convergence of (\ref{dvpr}) to zero. We show
that $z_{k-1}:={w_\mu\over w_k^\mu}= {\xi(y\!\pb
x_{<k})\over\mu(y\!\pb x_{<k})}\geq 0$ converges to a finite
value, which completes the proof. Let $\E$ denote the $\mu$
expectation. Then
\beqn
  \E[z_k|x_{<k}] =
  \sum_{x_k}\!'\mu(y\!x_{<k}y\!\pb x_k)
    {\xi(y\!\pb x_{1:k})\over\mu(y\!\pb x_{1:k})} =
  {\sum'_{x_k}\xi(y\!x_{<k}y\!\pb x_k)\xi(y\!\pb x_{<k})
    \over\mu(y\!\pb x_{<k})} \leq
  {\xi(y\!\pb x_{<k})\over\mu(y\!\pb x_{<k})} =
  z_{k-1}
\eeqn
$\sum'_{x_k}$ runs over all $x_k$ with $\mu(y\!\pb x_{1:k})\neq
0$. The first equality holds w.$\mu$.p.1. In the second equality
we have used Bayes rule twice. $\E[z_k|x_{<k}]\leq z_{k-1}$ shows
that $-z_k$ is a semi-martingale. Since $-z_k$ is non-positive,
\cite[Th.$4.1s(i)$,p324]{Doob:53} implies that $-z_k$ converges
for $k\to\infty$ to a finite value w.$\mu$.p.1. $\qed$

%%%%%%%%%%%%%%%%%%%%%%%%%%%%%%%%%%%%%%%%%%%%%%%%%%%%%%%%%%%%%%%
\section{Markov Decision Processes}\label{secMDP}
%%%%%%%%%%%%%%%%%%%%%%%%%%%%%%%%%%%%%%%%%%%%%%%%%%%%%%%%%%%%%%%

From all possible environments, Markov (decision) processes are
probably the most intensively studied ones. To give an example, we
apply Theorems \ref{thVcconv} and \ref{thVcconv2} to ergodic
Markov decision processes, but we will be very brief.

%------------------------------%
\fdefinition{defMDP}{Ergodic Markov Decision Processes}{
%------------------------------%
We call $\mu$ a (stationary) {\em Markov Decision Process ({\sc mdp})}
if the probability of observing $x_k\in\X$, given history
$y\!x_{<k}y_k$ does only depend on the last action $y_k\in\Y$ and
the last observation $x_{k-1}$, i.e.\ if $\mu(y\!x_{<k}y_k\pb
x_k)=\mu(y\!x_{k-1}\pb x_k)$. In this case $x_k$ is called a {\em
state}, $\X$ the {\em state space}, and $\mu(y\!x_{k-1}\pb x_k)$
the {\em transition matrix}. An {\sc mdp} $\mu$ is called {\em ergodic}
if there exists a policy under which every state is visited
infinitely often with probability 1. Let $\M_{MDP}$ be the set of
{\sc mdp}$s$ and $\M_{MDP1}$ be the set of ergodic {\sc mdp}$s$. If an {\sc mdp}
$\mu(y\!x_{k-1}\pb x_k)$ is independent of the action $y_{k-1}$ it
is a {\em Markov process}, if it is independent of the last
observation $x_{k-1}$ it is an {\em i.i.d.}\ process.
}%------------------------------%
Stationary {\sc mdp}$s$ $\mu$ have stationary optimal policies $p^\mu$
mapping the same state / observation $x_t$ always to the same action
$y_t$. On the other hand a mixture $\xi$ of {\sc mdp}$s$ is itself not an
{\sc mdp}, i.e.\ $\xi\not\in\M_{MDP}$, which implies that $p^\xi$ is, in
general, not a stationary policy.
The definition of ergodicity given here is least demanding, since
it only demands on the existence of a single policy under which the
Markov process is ergodic. Often, stronger assumptions, e.g.\ that
every policy is ergodic or that a stationary distribution exists,
are made.
We now show that there are self-optimizing policies for the class
of ergodic {\sc mdp}$s$ in the following sense.

%------------------------------%
\ftheorem{thMDPconv}{Self-optimizing policies for ergodic {\sc mdp}$s$}{
%------------------------------%
There exist self-optimizing policies $\tilde p_m$ for the
class of ergodic {\sc mdp}$s$ in the sense that
\beqn
i) \quad
  \exists\tilde p_m\forall\nu\!\in\!\M_{MDP1}:
 \odm V_{1m}^{\best\nu} - \odm V_{1m}^{\tilde p_m\nu}
 %\;=\; O(m^{-1/3})
 \;\leq\; c_\nu m^{-1/3} \;\toinfty{m}\; 0,
 \rule{35ex}{0ex}
\eeqn
where $c_\nu$ are some constants. In the discounted case, if the
discount sequence $\gamma_k$ has unbounded effective horizon
$h_k^{e\!f\!f}\toinfty{k}\infty$, then there exist self-optimizing
policies $\tilde\pi_k$ for the class of ergodic {\sc mdp}$s$ in
the sense that
\beqn
ii) \quad
  \exists\tilde \pi_k\forall\nu\!\in\!\M_{MDP1}:
  V_{k\gamma}^{\tilde \pi_k\nu} \;\toinfty{k}\; V_{k\gamma}^{\best\nu}
  \quad
  \mbox{if $\;{\gamma_{k+1}\over\gamma_k}\to 1$}.
  \rule{25ex}{0ex}
\eeqn\vspace{-3ex}
}%------------------------------%
There is much literature on constructing and analyzing
self-optimizing learning algorithms in {\sc mdp} environments. The
assumptions on the structure of the {\sc mdp}$s$ vary, all include some
form of ergodicity, often stronger than Definition
\ref{defMDP}, demanding that the Markov process is ergodic under
{\it every} policy. See, for instance,
\cite{Kumar:86,Bertsekas:95}. We will only briefly outline one
algorithm satisfying Theorem \ref{thMDPconv} without trying to
optimize performance.

\noindent{\bf Proof idea:} For $(i)$ one can choose a policy
$\tilde p_m$ which performs (uniformly) random actions in cycles
$1...k_0-1$ with $1\ll k_0\ll m$ and which follows thereafter the
optimal policy based on an estimate of the transition matrix
$T_{ss'}^a\equiv\nu(as\pb s')$ from the initial $k_0-1$ cycles.
The existence of an ergodic policy implies that for every pair of
states $s_{start}, s\in\X$ there is a sequence of actions and
transitions of length at most $|\X|-1$ such that state $s$ is
reached from state $s_{start}$. The probability that the ``right''
transition occurs is at least $T_{min}$ with $T_{min}$ being the
smallest non-zero transition probability in $T$. The probability
that a random action is the ``right'' action is at least
$|\Y|^{-1}$. So the probability of reaching a state $s$ in
$|\X|-1$ cycles via a random policy is at least
$(T_{min}/|\Y|)^{|\X|-1}$. In state $s$ action $a$ is taken with
probability $|\Y|^{-1}$ and leads to state $s'$ with probability
$T_{ss'}^a\geq T_{min}$. Hence, the expected number of transitions
$s\stackrel{a}\to s'$ to occur in the first $k_0$ cycles is $\geq
{k_0\over|\X|}(T_{min}/|\Y|)^{|\X|} \sim k_0$.\footnote{For
$T_{ss'}^a=0$ the estimate $\hat T_{ss'}^a=0$ is exact.} The
accuracy of the frequency estimate $\hat T_{ss'}^a$ of $T_{ss'}^a$
hence is $\sim k_0^{-1/2}$. Similar {\sc mdp}$s$ lead to
``similar'' optimal policies, which lead to similar values. More
precisely, one can show that $\hat T - T \sim k_0^{-1/2}$ implies
the same accuracy in the average value, i.e.\ $|\odm
V_{k_0m}^{\tilde p_m\nu} - \odm V_{k_0m}^{\best\nu}|\sim
k_0^{-1/2}$, where $\tilde p_m$ is the optimal policy based on
$\hat T$ and $\best$ is the optimal policy based on $T(=\nu)$.
Since $\odm V_{1k_0}\sim {k_0\over m}$, $(i)$ follows (with
probability 1) by setting $k_0\sim m^{2/3}$. The policy $\tilde
p_m$ can be derandomized, showing $(i)$ for sure.

The discounted case $(ii)$ can be proven similarly. The history
$y\!x_{<k}$ is simply ignored and the analogue to $m\to\infty$
is $h_k^{e\!f\!f}\to\infty$ for $k\to\infty$, which is ensured by
${\gamma_{k+1}\over\gamma_k}\to\infty$.
Let $\tilde \pi_k$ be the policy which performs (uniformly)
random actions in cycles $k...k_0-1$ with $k\ll k_0\ll h_k^{e\!f\!f}$
and which follows thereafter the optimal policy%
\footnote{For non-geometric discounts as here, optimal policies
are, in general, {\em not} stationary.} based on an estimate $\hat
T$ of the transition matrix $T$ from cycles $k...k_0-1$. The
existence of an ergodic policy, again, ensures that the expected
number of transitions $s\stackrel{a}\to s'$ occurring in cycles
$k...k_0-1$ is proportional to $\Delta:=k_0-k$. The accuracy of
the frequency estimate $\hat T$ of $T$ is $\sim \Delta^{-1/2}$
which implies
\beq\label{vtovk0}
  V_{k_0\gamma}^{\tilde \pi_k\nu} \to V_{k_0\gamma}^{\best\nu}
  \quad\mbox{for}\quad \Delta=k_0-k\to\infty,
\eeq
where $\tilde \pi_k$ is the optimal policy based on $\hat T$ and
$\best$ is the optimal policy based on $T(=\nu)$. It remains to
show that the achieved reward in the random phase $k...k_0-1$
gives a negligible contribution to $V_{k\gamma}$. The following
implications for $k\to\infty$ are easy to show:
\beqn
  {\gamma_{k+1}\over\gamma_k}\to 1
  \;\Rightarrow\;
  {\gamma_{k+\Delta}\over\gamma_k}\to 1
  \;\Rightarrow\;
  {\Gamma_{k+\Delta}\over\Gamma_k}\to 1
  \;\Rightarrow\;
  {1\over\Gamma_k}\sum_{i=k}^{k_0-1}\gamma_i r_i \leq
  {r_{max}\over\Gamma_k}[\Gamma_{k+\Delta}-\Gamma_k]\to 0.
\eeqn
Since convergence to zero is true for all fixed finite $\Delta$ it
is also true for sufficiently slowly increasing
$\Delta(k)\to\infty$. This shows that the contribution of the
first $\Delta$ rewards $r_k+...+r_{k_0-1}$ to $V_{k\gamma}$ is
negligible. Together with (\ref{vtovk0}) this shows
$V_{k\gamma}^{\tilde \pi_k\nu} \to V_{k\gamma}^{\best\nu}$ for
$k_0:=k+\Delta(k)$.$\qed$

The conditions $\Gamma_k<\infty$ and
${\gamma_{k+1}\over\gamma_k}\to 1$ on the discount sequence are,
for instance, satisfied for $\gamma_k=1/k^2$, so the Theorem is
not vacuous. The popular geometric discount $\gamma_k=\gamma^k$
fails the latter condition; it has finite effective horizon.
\cite{Hutter:00kcunai} gives a detailed account on discount and
horizon issues, and motivates $h_k^{e\!f\!f}\to\infty$
philosophically.

Together with Theorems \ref{thVcconv} and \ref{thVcconv2}, Theorem
\ref{thMDPconv} immediately implies that policy $p^\xi$ is self-optimizing
for the class of to ergodic {\sc mdp}$s$.

%------------------------------%
\fcorollary{coMDPconv}{Policy $p^\xi$ is self-optimizing for ergodic {\sc mdp}$s$}{
%------------------------------%
If $\M$ is a finite or countable class of ergodic {\sc mdp}$s$, and
$\xi():=\sum_{\nu\in\M}w_\nu \nu()$, then policies
$p_m^\xi$ maximizing $V_{1m}^{p\xi}$ and $p^\xi$ maximizing
$V_{k\gamma}^{\pi\xi}$ are self-optimizing in the sense that
\beqn
 \forall\nu\!\in\!\M:
 \odm V_{1m}^{p_m^\xi\nu} \;\toinfty{m}\; \odm V_{1m}^{\best\nu}
 \quad\mbox{and}\quad
  V_{k\gamma}^{p^\xi\nu} \;\toinfty{k}\; V_{k\gamma}^{\best\nu}
  \quad
  \mbox{if $\;{\gamma_{k+1}\over\gamma_k}\to 1$}.
\eeqn
If $\M$ is finite, then the speed of the first convergence
is at least $O(m^{-1/3})$.
}%------------------------------%

%%%%%%%%%%%%%%%%%%%%%%%%%%%%%%%%%%%%%%%%%%%%%%%%%%%%%%%%%%%%%%%
\section{Conclusions}\label{secConc}
%%%%%%%%%%%%%%%%%%%%%%%%%%%%%%%%%%%%%%%%%%%%%%%%%%%%%%%%%%%%%%%

%------------------------------%
\paragraph{Summary:}
%------------------------------%

We studied agents acting in general probabilistic environments
with reinforcement feedback. We only assumed that the true
environment $\mu$ belongs to a known class of environments $\M$,
but is otherwise unknown. We showed that the Bayes-optimal policy
$p^\xi$ based on the Bayes-mixture $\xi=\sum_{\nu\in\M}w_\nu \nu$
is Pareto-optimal and self-optimizing if $\M$ admits
self-optimizing policies. The class of ergodic {\sc mdp$s$}
admitted self-optimizing policies w.r.t.\ the average value and
w.r.t.\ the discounted value if the effective horizon grew
indefinitely.

%------------------------------%
\paragraph{Continuous classes $\M$:}
%------------------------------%
There are uncountably many (ergodic) {\sc mdp}$s$. Since we have
restricted our development to countable classes $\M$ we had to
give the Corollary for a countable subset of $\M_{MDP1}$. We may
choose $\M$ as the set of all ergodic {\sc mdp}$s$ with rational
(or computable) transition probabilities. In this case $\M$ is a
dense subset of $\M_{MDP1}$ which is, from a practical point of
view, sufficiently rich. On the other hand, it is possible to
extend the theory to continuously parameterized families of
environments $\mu_\theta$ and $\xi=\int
w_\theta\mu_\theta\,d\theta$. Under some mild (differentiability
and existence) conditions, most results of this work remain valid
in some form, especially Corollary \ref{coMDPconv} for {\em all}
ergodic {\sc mdp}$s$.

%------------------------------%
\paragraph{Bayesian self-optimizing policy:}
%------------------------------%
Policy $p^\xi$ with unbounded effective horizon for ergodic {\sc
mdp}$s$ is the first purely Bayesian self-optimizing consistent
policy for ergodic {\sc mdp}$s$. The policies of all previous
approaches were either hand crafted, like the ones in the proof of
Theorem \ref{thMDPconv}, or were Bayesian with a pre-chosen
horizon $m$, or with geometric discounting $\gamma$ with finite
effective horizon (which does not allow self-optimizing policies)
\cite{Kumar:86,Bertsekas:95}. The combined conditions
$\Gamma_k<\infty$ and ${\gamma_{k+1}\over\gamma_k}\to 1$ allow a
consistent self-optimizing Bayes-optimal policy based on mixtures.

%------------------------------%
\paragraph{Bandits:}
%------------------------------%
Bandits are a special subclass of ergodic {\sc mdp}$s$. In a
two-armed bandit problem you pull repeatedly one of two levers
resulting in a gain of A\$1 with probability $p_i$ for arm number
$i$. The game can be described as an {\sc mdp} with parameters
$p_i$. If the $p_i$ are unknown, Corollary \ref{coMDPconv} shows
that policy $p^\xi$ yields asymptotically optimal payoff. The
discounted unbounded horizon approach and result is, to the best
of our knowledge, even new when restricted to Bandits.

%------------------------------%
\paragraph{Other environmental classes:}
%------------------------------%
Bandits, i.i.d.\ processes, classification tasks, and many more
are all special (degenerate) cases of ergodic {\sc mdp}$s$, for
which Corollary \ref{coMDPconv} shows that $p^\xi$ is
self-optimizing. But the existence of self-optimizing policies is
not limited to (subclasses of ergodic) {\sc mdp}$s$. Certain
classes of {\sc pomdp}$s$, $k^{th}$ order ergodic {\sc mdp}$s$,
factorizable environments, repeated games, and prediction problems
are not {\sc mdp}$s$, but nevertheless admit self-optimizing
policies (to be shown elsewhere), and hence the corresponding
Bayes-optimal mixture policy $p^\xi$ is self-optimizing by
Theorems \ref{thVcconv} and \ref{thVcconv2}.

%------------------------------%
\paragraph{Outlook:}
%------------------------------%
Future research could be the derivation of non-asymptotic bounds,
possibly along the lines of \cite{Hutter:01loss}. To get good
bounds one may have to exploit extra properties of the
environments, like the mixing rate of {\sc mdp}$s$
\cite{Kearns:98}. Another possibility is to search for other
performance criteria along the lines of
\cite[Ch.6]{Hutter:00kcunai}, especially for the universal prior
\cite{Solomonoff:78} and for the Speed prior
\cite{Schmidhuber:02speed}. Finally, instead of convergence of
the expected reward sum, studying convergence with high
probability of the actual reward sum would be interesting.

%%%%%%%%%%%%%%%%%%%%%%%%%%%%%%%%%%%%%%%%%%%%%%%%%%%%%%%%%%%%%%%
%         Bibliography        %
%%%%%%%%%%%%%%%%%%%%%%%%%%%%%%%%%%%%%%%%%%%%%%%%%%%%%%%%%%%%%%%

\end{document}